\documentclass[letterpaper, 10 pt, conference]{ieeeconf}  

\IEEEoverridecommandlockouts                              

\title{\LARGE \bf
  A Mobile Manipulation System for One-Shot Teaching of Complex Tasks in Homes
}

\author{Max Bajracharya*, James Borders*, Dan Helmick*, Thomas
  Kollar* \\ Michael Laskey*, John Leichty*, Jeremy Ma*, Umashankar
  Nagarajan*, Akiyoshi Ochiai* \\ Josh Petersen*, Krishna Shankar*,
  Kevin Stone*, Yutaka Takaoka* \\ Toyota Research Institute, Los
  Altos, CA \\ {\tt\small *all authors contributed equally.}
}

\usepackage{graphicx} 
\usepackage{mathptmx} 
\usepackage{times} 
\usepackage{amsmath} 
\usepackage{amssymb}  
\usepackage{algorithm}
\usepackage{algorithmic}
\usepackage{color}
\usepackage{hyperref}

\newcommand{\degree}{\ensuremath{^\circ}}

\begin{document}

\maketitle
\thispagestyle{empty}
\pagestyle{empty}

\bstctlcite{icra-ref:BSTcontrol}

\begin{abstract}
  We describe a mobile manipulation hardware and software system capable of autonomously performing complex human-level tasks in real homes, after being taught the task with a single demonstration from a person in virtual reality.  This is enabled by a highly capable mobile manipulation robot, whole-body task space hybrid position/force control, teaching of parameterized primitives linked to a robust learned dense visual embeddings representation of the scene, and a task graph of the taught behaviors.  We demonstrate the robustness of the approach by presenting results for performing a variety of tasks, under different environmental conditions, in multiple real homes.  Our approach achieves 85\% overall success rate on three tasks that consist of an average of 45 behaviors each.  The video is available at: \href{https://youtu.be/HSyAGMGikLk}{https://youtu.be/HSyAGMGikLk}.
\end{abstract}


\section{INTRODUCTION}
\label{section:introduction}


Robotic capabilities that assist people with tasks in their homes can play a critical role in enabling them to age in place longer and live a higher quality life.  However, the tasks people perform in their homes vary widely, 
and home environments, objects, and tasks are highly unstructured and extremely diverse.
But one advantage a robot operating in a home has is that it only needs to work well in that home, and its actions can be specialized to that environment.

Based on these observations, we have developed a unique solution to enabling a general purpose robot to perform human-level tasks in diverse and complex human environments.  
Rather than program or train a robot to recognize a fixed set of objects or perform pre-defined tasks, we enable the robot to be easily taught new objects and tasks, with inherently robust behaviors, from a single human demonstration, which can then be executed autonomously in naturally varying conditions.  Our system uses no prior object models or maps, and can be taught to associate a given set of behaviors to arbitrary scenes, objects, and voice commands from one demonstration of the behavior.  Because tasks are graphs of behaviors linked to dense visual features, the system is easy to understand and failure conditions are easy to diagnose and reproduce.


\begin{figure}[t!]
  \centering
  \includegraphics[width=0.85\linewidth]{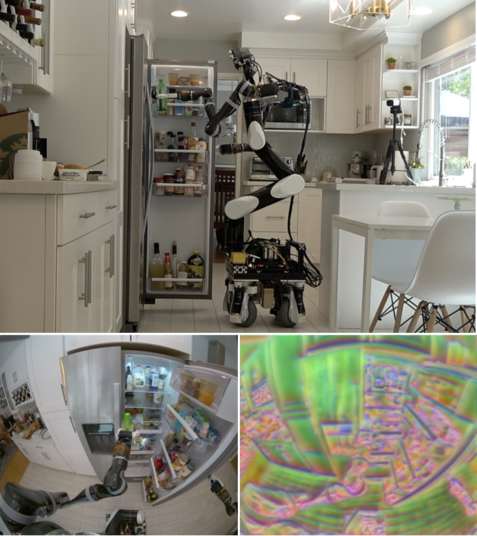}
  \caption{We have developed a highly capable general purpose mobile manipulation robot (top) capable of being taught human-level tasks by linking what it sees (bottom left) to robust parameterized behaviors, through dense learned visual features (whose first three dimensions are shown on the bottom right), from a single human demonstration of the behavior.}
  \label{fig:setup}
\end{figure}

Our solution consists of several key contributions:
\begin{enumerate}
\item We developed a mobile manipulation robot that is very physically capable, with high end-effector manipulability and a wide instantaneous visual field-of-view, which makes teaching from human demonstration easy.
\item Rather than teach direct task space motions, we use virtual reality (VR) to teach a set of parameterized behaviors, which combine collision-free motion planning and whole-body hybrid (position, velocity, admittance) Cartesian end-effector control, minimizing the taught parameters and ensuring robustness during execution.
\item We compute task specific learned dense visual pixelwise embeddings (keyframes), which link the parameterized behaviors to the scene and enable it to compute a 6-DOF transform of taught behaviors at execution time.  While robust to viewpoint change, lighting variations, and light clutter, we do not attempt to generalize beyond specific taught situations.
\item The behaviors of a task are taught independently, with visual entry conditions, and success-based exit criteria, which enable behaviors to be chained together in a dynamic task graph, allowing the robot to reuse taught behaviors to perform task sequences.
\end{enumerate}

Our approach applies to both manipulation tasks, as well as navigation, where a taught trajectory is identified by keyframes along the path, and followed with a reactive path follower.  Since actions are taken with respect to taught keyframes, an explicit global map is not used for either navigation or manipulation.
The approach achieves a success rate of 85\% on a set of challenging tasks that have an average of 45 behaviors per task.
The use of learned dense embeddings makes the system robust to expected daily changes in the environment, including lighting and clutter.  The use of parameterized hybrid control behaviors makes the system robust to limited accuracy mechanisms.




\subsection{Related Work}
\label{section:relatedowrk}

There are many research mobile manipulation robots
including those
with a wheeled base, linear stage torso, and a single arm \cite{robotnik}, \cite{fetch}, \cite{hsr} or dual arms \cite{pr2}, an arm on a quadruped legged base \cite{spot}, or a more humanoid form factor on wheels \cite{halodi} or legs \cite{atlas}, among others.  None that can be acquired provide a form factor that allow for performing tasks in real homes.

The 2015 DARPA Robotics Challenge Finals \cite{drc} presented a snapshot of many fully integrated mobile manipulation systems being remotely operated with semi-autonomous capabilities.  Prior to that, the DARPA Autonomous Robotic Manipulation Software (ARM-S) program demonstrated performing complex dual-arm manipulation tasks fully autonomously \cite{hudson13modelbased}, \cite{righetti14autonomous}, \cite{bagnell12integrated}.  However, these approaches relied heavily on prior models of objects and took significant engineering effort to perform new tasks.

Some core elements of our approach are commonly used in industrial manipulation applications, such as parameterized behaviors \cite{kroger10manipulation}, hybrid position-force control \cite{backes91generalized} \cite{kroger10hybrid}, task graph formulations including Petri nets and hierarchical state machines, and kinematic teach and repeat, but have been limited to structured and controlled environments.

Visual teach and repeat has been used in less structured environments for outdoor ground navigation \cite{furgale10visual} and aerial localization \cite{fehr18visualinertial}.  For manipulation, a related approach is trajectory transfer \cite{schulman2016learning}, where pixel-level correspondences between a reference scene and the current environment are used to warp a taught sequence to a new initial state.  Unlike this approach, we apply registration at the behavioral level as opposed to the whole trajectory.  We extend the idea to associate behaviors using generic learned dense pixelwise visual embeddings for feature matching \cite{schmidt2016self}, rather than estimating constraints from semantic class specific sparse keypoints \cite{florence18dense}.

Imitation learning from VR demonstration can also learn a mapping of images to control, such as using behavioral cloning \cite{zhang2018deep}, but still requires a large amount of data. Approaches to single-shot learning from demonstration include variants of model-based and model-free inverse reinforcement learning, online reinforcement learning, and meta learning, \cite{englert18learning}, \cite{fu15oneshot}, \cite{finn17oneshot}, \cite{zhou19watch} but these all still require significant amounts of data for training similar tasks, or a large number of policy executions, before being able to perform well on a new task.  Some approaches also automatically generate the task graph \cite{huang18neural}.
Training behaviors in simulation and transfering to reality has shown some promise \cite{chebotar18closing} but still requires some online refinement and is limited to what can be simulated.

\section{HARDWARE SYSTEM}
\label{section:hardware}

We have developed a custom prototype mobile manipulation robot physically capable of performing a wide variety of household tasks, and specifically designed to make teaching of tasks easy.
Making the system low cost was not a priority, but recent advances in sensors and actuators \cite{gealy19quasi} \cite{halodi} indicate that doing so is possible, and our approach is designed to be robust to low precision mechanisms.  The choice of components was based on making the overall system lightweight, compact, and power efficient enough to perform tasks in real homes.
We have experimentally found that a person in VR, seeing only what the robot sees and controlling its end-effectors, can do many household tasks, except highly dexterous or very high payload tasks.


\subsubsection{Morphology and Actuation}
The 100kg robot consists of a total of 31 degrees-of-freedom (DOFs).
The chassis consists of four driven and steerable wheels (eight total DOFs) that enable ``pseudo-holonomic'' mobility.  The drive/steer actuator package is a custom modular design using brushless Maxon motors and planetary gearheads.  The torso is five DOFs (yaw-pitch-pitch-pitch-yaw) built using a Motiv Robotics RoboMantis limb, which is derived from the JPL RoboSimian limb \cite{robosimian}.  Each arm is a seven DOF Kinova Jaco2 arm.  The two DOF pan/tilt head also uses Kinova actuators.
Each arm also has a single DOF Sake Robotics gripper with under-actuated fingers.  We modified the Sake gripper fingers to have 3D printed hooks that greatly improve the ability to pull handles or knobs with high force, without any observed impact on snagging in high clutter situations or other negative side effects.  We can also manually replace the gripper with custom tools, such as a sponge or a wiping pad, to enable different tasks.  




\subsubsection{Sensing}
We use an ATI mini-45 force/torque sensor
at the wrist of each arm to measure interaction forces with the environment.  Our perception sensors are consolidated on the pan/tilt head of the robot, with a very wide field-of-view, giving the robot and a person in VR significant context to perform the task.  They consist of four Intel RealSense depth cameras, a pair of 5MP Basler cameras with a 7cm baseline, and a VectorNav IMU.  The RealSense cameras are arranged in a 2x2 configuration to produce a depth image with a total field-of-view (FOV) of 110\degree x80\degree and the Baslers each have an FOV of 146\degree x123\degree.  The cameras use USB 3.0
and are hardware triggered and timestamped by the computer.  The six cameras are calibrated using Kalibr~\cite{kalibr}, with a double-sphere fisheye model~\cite{usenko18double-sphere} for the stereo pair.

\subsubsection{Compute}
All computation is performed on-board the robot.  The compute system consists of an 18 core Intel i9 CPU and an Nvidia TitanV GPU.  We use a Linux kernel with the Preempt RT patch applied.  This compute system allows us to use a single computer for all of our processes, including both real-time control and perception.  All of our inference is done on the GPU using TensorRT models.

\subsubsection{Power}
We use six standard BB2590 Li-Ion batteries, each with 294Wh of capacity.  While running and performing tasks, the system draws between 650W and 750W.  We developed a custom power board that handles power distribution, on-board battery charging, and emergency stopping.




\section{SOFTWARE SYSTEM}
\label{section:system_architecture}

\begin{figure}
  \vspace*{0.2cm}
  \centering
  \includegraphics[width=0.95\linewidth]{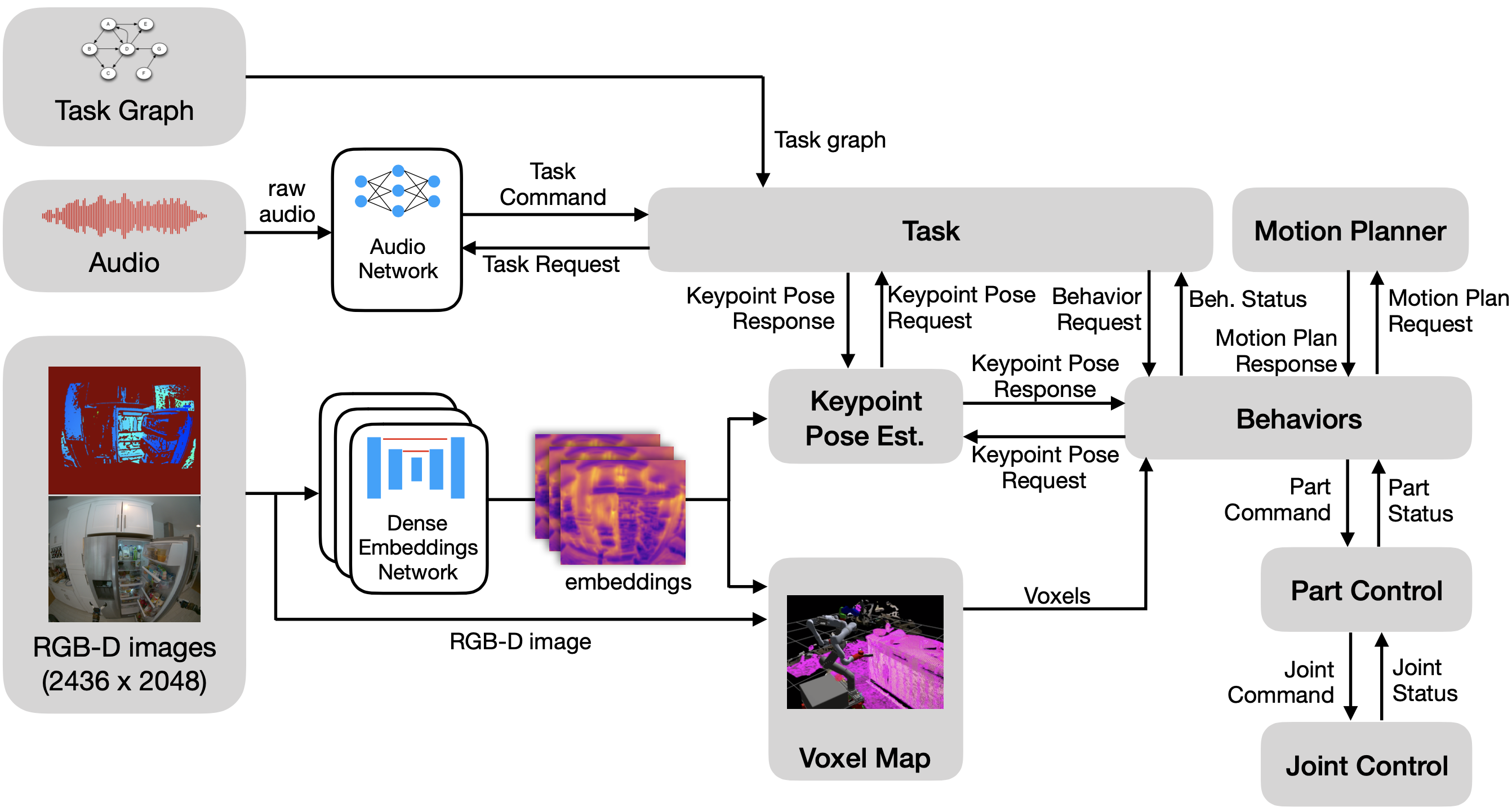}
  \caption{Our software architecture enables robust autonomous execution of taught tasks by processing visual and audio data, building up a world model, mapping visual inputs to taught behaviors, and executing sequences of behaviors.}
  \label{architecture}
\end{figure}

Our software system (Figure \ref{architecture}) is designed to leverage our highly redundant and capable hardware safely and effectively.  It is specifically architected to enable teaching of behaviors.  It is also designed to enable fast iterative development and deployment, debugging, and visualization of the system.

\subsection{Software Architecture}
\subsubsection{Infrastructure}

\label{section:infrastructure}

The system is architected like many standard robotic systems \cite{reeves05overview} \cite{quigley09ros}, with independent processes communicating via messages over a custom interprocess communication (IPC) implementation.
The system is organized as a set of modules, one or more of which run in a system process, which handle a set of input messages and publish a set of output messages.  All messages are logged and modules or sets of modules can be replayed deterministically in a single process, or as if it were running on the robot, in parallel.


\subsubsection{Visualization, Commanding, and Teaching}
\label{section:vr_overview}

We developed a custom visualization tool that subscribes to messages and can display 2D, 3D, text, and temporal information.  It can also be used to command the robot and inspect and modify the task sequences that the robot is capable of executing.
The robot state and RGB-D data is also streamed live into a VR system, enabling a person to teach the robot parameters of behaviors.
We define these parameters by using an
HTC Vive VR headset with two hand
controllers. This allows an operator to see the world from the
perspective of the robot, annotate the 3D point cloud,
and float detached robot end-effectors in space to
define end-effector poses.


\subsection{Control Architecture}
Our system provides several key levels of abstraction for controlling the robot, specifically making it easy to teach and execute robust task sequences.

\subsubsection{Real-time Control}

\label{section:realtimecontrol}

The lowest levels provide real-time coordinated control of all of the robot's DOFs.  Real-time control consists of two processes working in coordination at 200Hz:  Joint Control and Part Control.
Joint Control implements low-level device communications and 
exposes the device commands and statuses in a generic way.
It also provides the lowest level of safety checks. If an 
incoming command violates center of mass constraints, causes a 
self-collision, or violates joint state limits, a fault is triggered and the 
robot is brought to rest safely.


Part Control handles higher level coordination of the robot by dividing the 
robot into parts (right arm, head, etc.) and providing a set of parameterized 
controllers for each part. Commands from non-realtime processes set the desired 
controllers and parameters to be running at a given time. Arbitrary 
combinations of controllers are supported as long as their controlled parts do 
not overlap. It provides controllers for joint position and velocity, joint admittance, camera look-at, chassis position and 
velocity, and hybrid task space pose, velocity, and admittance control.



\subsubsection{Whole-body Planning}
\label{section:whole_body_planning}

The next level of abstraction for controlling the robot is commanding end-effector task space control and automatically solving for the robot posture to achieve these desired motions.
Whole-body inverse kinematics (IK) for hybrid Cartesian control are formulated 
as a quadratic program (QP) \cite{QP} and solved in real-time at 
200Hz. Parts are subject to linear constraints on joint position, velocity, 
acceleration, and torque due to gravity, center of mass, and self-collisions and 
quadratic costs on Cartesian tracking, regularization, and distance to 
preferred postures.



Whole-body IK are used for non-realtime motion 
planning of Cartesian pose goals. Occupied environment voxels 
(Section \ref{section:voxel_mapper}) are fit with spheres and capsules and voxel 
collision constraints are added to the QP IK to prevent collisions between the 
robot and the world. Motion planning is performed using a rapidly-exploring 
random tree (RRT) \cite{RRT}, sampling in Cartesian space with the QP IK as the 
steering function between nodes. Planning in Cartesian space results in natural and direct motions, and using the QP IK as the steering function makes 
planning more reliable, as the same controller is used to plan and execute, 
reducing the possible discrepancies between the two.



\subsubsection{Parameterized Behaviors}
\label{section:behaviors}

The next level of abstraction defines parameterized behaviors, which are primitive actions that can be parameterized and sequenced together to accomplish a complex task.
We have found that a small set of parameterized behaviors are sufficient to perform many tasks, however the software architecture supports quick addition of new behaviors as and when they are necessary. Our behaviors include (1) manipulation actions such as grasp, lift, place, pull, retract, wipe, joint-move, direct-control; (2) navigation actions such as drive with velocity commands, drive-to with position commands and follow-path with active obstacle avoidance~\cite{path_follower}; and (3) other auxiliary actions such as look at and stop.

Each behavior can have single or multiple actions of different types such as joint or Cartesian moves for one or more parts of the robot. 
Each action can use different control strategies such as position, velocity or admittance control, and can also choose to use motion planning to avoid external obstacles or not. 
All motions, whether they use motion planning or not, ensure that there is no self-collision and that all motion control constraints are satisfied. 
Each behavior is parameterized by the different actions, which in turn will have their own parameters.
For example, a grasp behavior consists of four parameters: gripper open angle, 6D approach, grasp and (optional) lift poses for the gripper. These four parameters define the following pre-defined sequence of actions: (1) open the gripper to desired gripper angle, (2) plan and execute a collision-free path for the gripper to the 6D approach pose, (3) move the gripper to the 6D grasp pose and stop on contact, (4) close the gripper, and (5) move the gripper to the 6D lift pose, if provided.


\subsubsection{Task Graphs}
\label{section:taskgraphs}

The final level of control abstraction is a task.  A task is a sequence of sub-tasks made up of taught behaviors. A task graph is a directed, cyclic or acyclic graph with different sub-tasks as nodes and different transition conditions as edges, including fault detection and fault recovery.  Edge conditions include the status of each behavior execution, checking for objects in hand using force/torque sensors, voice commands, and keyframe matches to handle different objects and environments.  The task graph is created at teach time by manually specifying nodes and transitions.

\subsection{Perception Architecture}
Our perception pipeline is designed to provide the robot with an understanding of the environment around it and to recognize what actions to take, given the task it has been taught.  A single fused RGB-D image is created by projecting the four depth images into the wide field-of-view left image of the high resolution color stereo pair.  The system runs a set of deep neural networks to provide various pixel level classifications and feature vectors (or ``embeddings'') which are then both accumulated into a temporal 3D voxel representation (Figure \ref{architecture}), as well as used to recall actions to perform, based on the visual features from a taught sequence.

\subsubsection{Learned Dense Pixel Embeddings}

\label{section:inference}

Based on experience testing in highly unstructured and diverse (``long tailed'' \cite{horn17devil}) environments, like homes, a key aspect of our system is that we do not pre-define object categories or assume any models of objects or the environment.  Rather than explicitly detect and segment objects \cite{he17mask}, and explicitly estimate 6-DOF object poses \cite{tremblay18deep}, we instead produce dense pixel level embeddings for object semantic classes and instances, and viewpoint invariant correspondences, and use the reference embeddings from a taught reference behavior to perform classification or pose estimation.

All of our learned models are fully convolutional, and map every pixel
in the input RGB image to a point in an embedding space with a metric that is implicitly defined by a loss function and training procedure specific to each model.  All models use a common feature extractor, which
consists of a ResNet \cite{Szegedy:2017} 101
encoder, and a variant of the Feature-Pyramid Network
\cite{Lin:2017} decoder. Given an input RGB image of size
$\textrm{height} \times \textrm{width} \times 3$, the feature
extractor produces an output of size $\frac{1}{8}\textrm{height} \times
\frac{1}{8}\textrm{width} \times 2048$,
which is fed to a final $1 \times 1$ convolution with an output depth that is
chosen depending on the output.  We use models trained for:
\begin{itemize}
\item{Semantic class: We detect all objects of a semantic class
  given a single annotated example
  by comparing the embeddings on the annotation to the
  embeddings we see everywhere else. We train this model using a
  discriminative loss function as described in \cite{Vangool:2017},
  on the MSCOCO data set \cite{lin2014microsoft}.}
\item{Object instance: This model is necessary for identifying or
  counting individual objects.
  We train the model to predict a vector (2D embedding)
  at each pixel, pointing to the centroid of the object containing
  that pixel. At run-time, we group all pixels that point to the same
  centroid to
  segment the scene.}
\item{3D correspondence: This model produces per pixel embeddings that are
  invariant to view and lighting, so that any view of a given 3D point
  in a scene will map to the same embedding.
  We train this model using
  the same approach and loss function described in
  \cite{schmidt2016self}, on the ScanNet data set
  \cite{dai2017scannet}}.
\end{itemize}

All of our models are written in TensorFlow, and converted and run
on-board the robot with Nvidia TensorRT using 16-bit floating point
precision on an Nvidia Titan-V GPU, with multiple processes coordinated using Nvidia's Multi-Process Service (MPS),
achieving about 90 megapixels per second.


\subsubsection{Voxel Mapping}
\label{section:voxel_mapper}

The pixelwise embeddings (and depth data) for each RGB-D frame is then fused into a dynamic 3D voxel map  \cite{bajracharya2012realtime}.  Each voxel accumulates first and second order position, color, and embeddings statistics.  Expiration of dynamic objects is based on back projection of voxels into the depth image.  The voxel map is segmented using standard graph segmentation based on the semantic and instance labels, and geometric proximity.  The voxel map is also collapsed down into a 2.5D map with elevation and traversability classification statistics.

The voxel map is used for collision free whole-body motion planning, while the 2.5D map is used for collision free chassis motions.
The segmented objects are used by the behaviors to attach objects to hands when they are grasped.


\subsubsection{Keypoint Pose Estimation}
\label{section:keyframe_matching}

\begin{figure}
  \vspace*{0.2cm}
  \centering
  \includegraphics[width=0.95\linewidth]{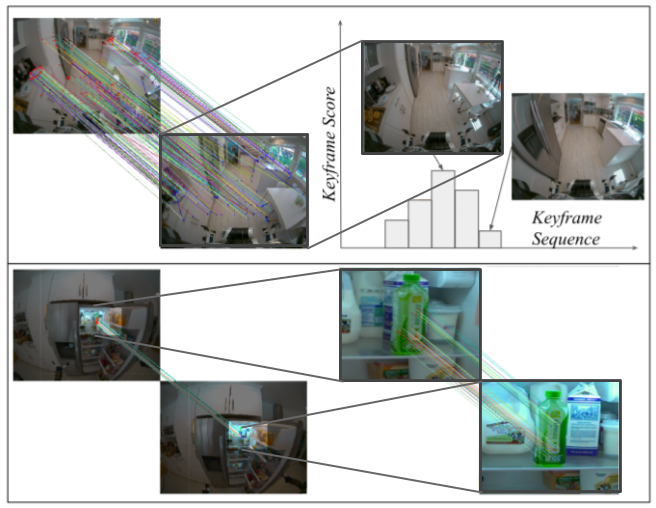}
  \caption{We use dense learned embeddings and geometric constraints to match a current scene (top) or part of a scene (bottom) to a previously taught one.  For behavior sequences or various entry conditions, the best keyframe is computed and selected from a set (top right).}
  \label{keyframe_matching}
\end{figure}

Central to our one-shot teaching approach is being able to recognize features in the
scene (or of a specific manipulation object) that are highly correlated to
features recorded from a previously taught task.
When a task is demonstrated by the user,
features are saved throughout the task in the form of a keyframe,
a saved RGB image containing a multi-dimensional embedding with depth
(if valid) per pixel. The embeddings act as a feature descriptor that
is ideally unique enough to establish per pixel correspondences at
run-time, assuming that the current image or object is visually similar enough to the
reference that existed at teach time. The keypoints are trained to be viewpoint invariant, but not semantically meaningful, and tend to latch onto specific textures or visual features.  
Since depth exists at (mostly)
each pixel, correspondences can be used to solve for a delta pose
between the current and reference images.  Our keyframe matcher detects inliers using Euclidian constraints \cite{hirschmuller03fast} and applies the
standard Levenberg-Marquardt least-squares algorithm with RANSAC to solve for
a 6-DOF pose. This delta pose serves as a correction that can be applied to adapt
the taught behavior sequence to the current
scene. Because we have
embeddings defined at each pixel, we can
define keyframes including every pixel in the image or only using pixels in a user-defined mask (where we
selectively annotate regions of the image to be relevant for the task) or
on an object (Figure \ref{keyframe_matching}).
Our approach also allows for multiple keyframes to be
passed to the matching problem,
which chooses the best keyframe at run-time based on the number of
correspondences found.

\subsubsection{Audio Processing}
\label{section:voice_command}

In addition to visual sensing, we also collect and process audio input.  Ultimately, the audio provides another set of embeddings as input for teaching the robot, but for now we only train the system to recognize specific spoken words.  The robot acquires input by asking questions and understanding spoken language responses from a person.

Spoken questions are produced using the eSpeak synthesizer.  Spoken
responses are understood using a custom keyword-detection module.  The
robot can understand a custom wakeword, a set of objects (e.g.,
``mug'' or ``bottle'') and a set of locations (e.g., ``cabinet'' or
``fridge'') using a fully-convolutional keyword-spotting model.  The
input to the model is single-channel 16-bit audio captured at 16 kHz,
from which a spectrogram and MFCC features are extracted.  The input
audio clip duration is 1300 ms, the spectrogram window is 30 ms, and the
number of MFCC bins is 40.  The model is trained using cross-entropy
loss, and consists of three layers of convolutions, with max pooling
after each layer, and ReLU activation functions.


The model listens for
the wakeword every 32 ms; when the wakeword is detected, it looks to
detect an object or location keyword in the following 2500 ms.  A
keyword must have been detected at least three times with at least
probability 0.5 in order to be recognized.
During training, noise is
artificially added to make recognition more robust.  The offline
accuracy at identifying individual keywords is 98\%.

\subsubsection{Teaching}
\label{section:teach_repeat}

\begin{figure}
  \vspace*{0.2cm}
  \centering
  \includegraphics[width=0.85\linewidth]{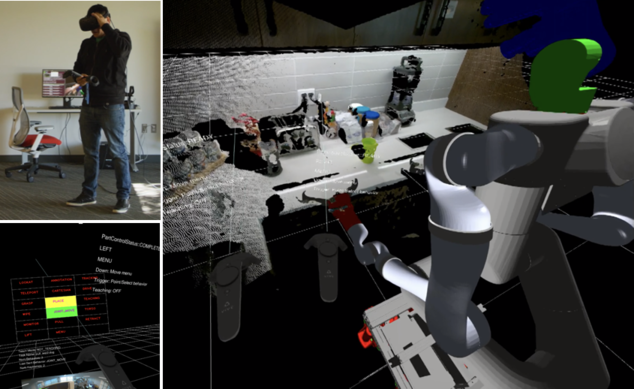}
  \caption{A person in virtual reality (upper left) can teach a variety of parameterized behaviors (shown in the menu on the lower left) by visualizing a robot model, what the robot is seeing, and flying tools around in 3D (right) to define the parameters of the behavior.}
  \label{teaching}
\end{figure}

To teach the robot a task, the operator uses a set of VR modes (Figure \ref{teaching}). Each behavior 
has a corresponding VR mode for setting and commanding
the specific parameters of that behavior. Each behavior mode has customized 
visualizations to aid in setting each parameter, dependent on the type of 
parameter. For example, when setting the parameters for a pull door motion, the 
hinge axis is labeled and visualized as a line and candidate pull poses for the 
gripper are restricted to fall on the arc about the hinge.
When appropriate, force vector setpoints are specified during teaching to define interaction forces desired during execution of the behavior.
\section{EXPERIMENTAL RESULTS}
\label{section:experimental_results}

To evaluate the robustness of our system and approach, experiments were performed with the mobile manipulation robot in multiple real homes.  Here we present three tasks, performed ten times, in two homes for a total of 60 experiments in order to obtain a measurement of task robustness across natural variations (e.g. lighting conditions during different times of the day, minor variations in initial object poses, wheel slippage,
etc.).  No software or parameter value changes were made across any of the 60 experiments, and each task was taught only once for the 10 experiments done for each task in each home.  The homes were not modified in any way, except for removing personally identifiable information from the scene.  The robot operated entirely autonomously for each of the 60 experiments.  Additionally, ad hoc experiments were performed with intentional variations of the home in order to test the robustness of the system.  

\subsection{Task Descriptions}
\label{section:task_descriptions}

The three tasks that we evaluated were:

\subsubsection{Task 1: Bottle from Refrigerator}
The robot starts in a different room than the kitchen, drives to the kitchen, opens the refrigerator, grasps a bottle, closes the refrigerator, and then drives back to the original room with the bottle.  The experiment is considered a success if the robot returns to the original location with the bottle.


\subsubsection{Task 2: Cup from Dishwasher}
The robot opens the dishwasher, removes a cup, closes the dishwasher, and places the cup on the countertop.  The experiment is considered a success if the cup ends on the countertop.


\subsubsection{Task 3: Moving Multiple Objects to Multiple Locations}
The robot asks the user which object to put away, grasps that object, asks the user where to put the object, then drives to the specified location and puts the object away.  In these experiments, we used two objects (a cup and a bottle) and two locations (a table and a cabinet).  Voice commands were used to specify the object and location.  Additionally, the cabinet door could be in any of three states: open, closed, or partially open.
The experiment is considered a success if the object ends in the specified location.

In addition to the 60 experiments performed for measurement of natural variation robustness, several more experiments for each task were performed to test the bounds of the robustness of the system using intentional variations.  Task~1 variations included: putting the bottle on a different shelf than it was taught on, adding obstacles along the path, adding pictures/magnets to the refrigerator, and varying the lighting conditions by closing blinds and turning lights on/off.  Task~2 variations included: varying the lighting conditions.  Task~3 variations included: swapping the initial positions of the two objects, opening adjacent cabinet doors, re-arranging the items in the cabinet, adding obstacles along the path, and adding a placemat to the table.


\subsection{Task Results}
\label{section:task_results}

\begin{figure}
  \vspace*{0.2cm}
  \includegraphics[width=1.0\linewidth]{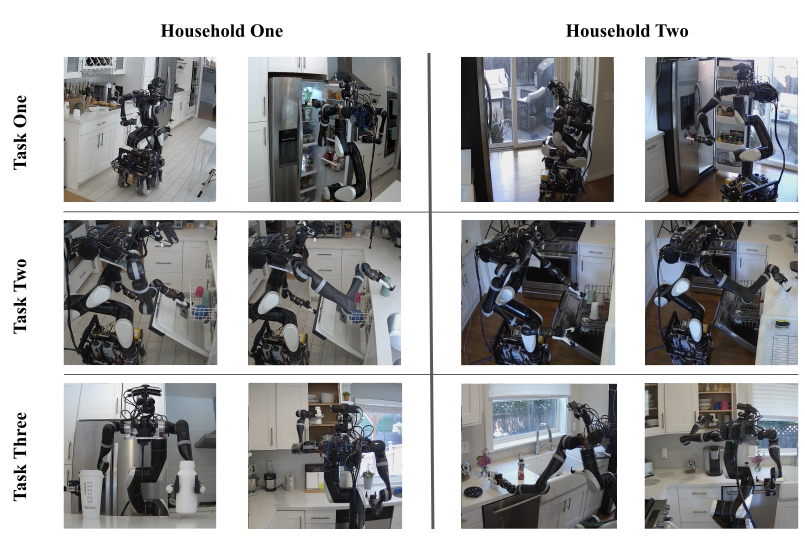}
  \caption{We performed a variety of tasks in multiple homes to evaluate the robustness of our system.  The images here show autonomous execution of parts of the tasks in different homes.}
\label{tasks}
\end{figure}

Of the 60 end-to-end task experiments perfomed in two homes (with examples shown in Figure \ref{tasks}), 51 were successful in completing the task, resulting in an overall success rate across all three tasks and both homes of 85\%.  A significant contributor to the end to end success rate of our tasks was fault detection and recovery within the task graph.
On average, our three tasks consisted of 45 behaviors each that are executed in series.  This means that behaviors result in success or recoverable failure 99.6\% of the time (or irrecoverable failure 0.4\% of the time).  For the three tasks, the robot performs the task anywhere from 10x to 100x slower than a person performing the same task, with the average being 20x slower.





The task failures are all the result of two different failure modes.  The first is that the pose estimate of the object or affordance is inaccurate, resulting in the behavior positioning the end-effector in such a way that the behavior fails (e.g. the gripper slips off the handle).  The second is that the scene appears to look too much like a different keyframe than the desired behavior (e.g. the partial open cabinet appears to be closed), and so the wrong behavior is performed.  For these experiments, no task failure was catastrophic, so if the robot had the error detection required for the failed cases, it could have tried again and potentially succeeded at the task.

The system is quite robust to intentional variations of the scene and task.  For example, the robot was able to successfully grasp the bottle from a different shelf in the refrigerator than it was taught on, it was able to avoid obstacles placed along the taught paths, and the keyframe matcher showed robustness to significant lighting changes and environmental changes such as opening cabinet doors and adding pictures to the refrigerator.  There were a few systematic failures that were found with these intentional variations, such as large rotations of the objects to be grasped.





\section{CONCLUSION}
\label{section:conclusion}

The combination of a highly capable and manipulable mobile robot with the ability to teach robust parameterized behaviors linked to dense visual embeddings from human demonstration in VR has proven to be surprisingly effective and robust to performing a wide variety of human-level tasks in real homes.  While not able to generalize beyond the taught scenario, tasks are tolerant to natural variation that occurs in home environments.
Because perception and behaviors are cleanly decoupled, much of the system could be tested and evaluated (or even synthesized) in simulation, which is likely key to eventually scaling a system to real users.

A key limitation of the current approach is that it requires teaching every task in VR, including explicitly annotating relevant parts of the scene, such as objects or articulated regions, for all possible discrete states of the environment (e.g. cabinet door open versus closed).  Incremental improvements, such as automatically determining the relevant parts of the scene (based on what the robot picks up or moves, for example), and teaching with multiple views of a scene or object, could help alleviate some of these limitations.







\end{document}